\documentclass[letterpaper]{article} 
\usepackage{aaai24}  
\usepackage{times}  
\usepackage{helvet}  
\usepackage{courier}  
\usepackage[hyphens]{url}  
\usepackage{graphicx} 
\usepackage{natbib}  
\usepackage{caption} 
\usepackage{algorithm}
\usepackage[algo2e, ruled]{algorithm2e}

\usepackage{xcolor}
\usepackage{url}
\usepackage{pdfpages}
\usepackage{amssymb}
\usepackage{pifont}
\newcommand{\cmark}{\ding{51}}%
\newcommand{\xmark}{\ding{55}}%
\usepackage{amsmath}

\usepackage{newfloat}
\usepackage{listings}
\DeclareCaptionStyle{ruled}{labelfont=normalfont,labelsep=colon,strut=off} 
\floatstyle{ruled}
\newfloat{listing}{tb}{lst}{}
\floatname{listing}{Listing}
\title{Descanning: From Scanned to the Original Images with \\ a Color Correction Diffusion Model}
\author{
    Junghun Cha\textsuperscript{\rm 1}\equalcontrib,
    Ali Haider\textsuperscript{\rm 1}\equalcontrib,
    Seoyun Yang\textsuperscript{\rm 1},
    Hoeyeong Jin\textsuperscript{\rm 1},
    Subin Yang\textsuperscript{\rm 1},\\
    A. F. M. Shahab Uddin\textsuperscript{\rm 2}, 
    Jaehyoung Kim\textsuperscript{\rm 1},
    Soo Ye Kim\textsuperscript{\rm 3},
    Sung-Ho Bae\textsuperscript{\rm 1}
}
\affiliations{
    \textsuperscript{\rm 1} Kyung Hee University\\
    \textsuperscript{\rm 2} Jashore University of Science and Technology \\
    \textsuperscript{\rm 3} Adobe Research\\
    \{jhcha08, alihaider, 2019101234, wlsghldud, ysb8049, crux153, shbae\}@khu.ac.kr, s.uddin@just.edu.bd, sooyek@adobe.com

}

\begin{document}

\maketitle

\begin{abstract}
A significant volume of analog information, i.e., documents and images, have been digitized in the form of scanned copies for storing, sharing, and/or analyzing in the digital world. However, the quality of such contents is severely degraded by various distortions caused by printing, storing, and scanning processes in the physical world. Although restoring high-quality content from scanned copies has become an indispensable task for many products, it has not been systematically explored, and to the best of our knowledge, no public datasets are available. In this paper, we define this problem as \textbf{Descanning} and introduce a new high-quality and large-scale dataset named \textbf{DESCAN-18K}. It contains 18K pairs of original and scanned images collected in the wild containing multiple complex degradations. In order to eliminate such complex degradations, we propose a new image restoration model called \textbf{DescanDiffusion} consisting of a color encoder that corrects the global color degradation and a conditional denoising diffusion probabilistic model (DDPM) that removes local degradations. To further improve the generalization ability of DescanDiffusion, we also design a synthetic data generation scheme by reproducing prominent degradations in scanned images. We demonstrate that our DescanDiffusion outperforms other baselines including commercial restoration products, objectively and subjectively, via comprehensive experiments and analyses.
\end{abstract}

\section{Introduction}

In the last several decades, information in the form of general paper-type materials (e.g., magazines, books, or photos) has been actively digitized via scanning processes, to store, share and analyze such information in digital form.
For instance, Google has scanned and digitized more than 25 million books under the codename Project Ocean \cite{ocean} since 2002.
However, the quality of scanned images is often degraded due to the printing, storing, and scanning processes.
Thus, to preserve the original information accurately, degradations caused by such processes should be removed from the digitized (scanned) copies.
Technically, as each scanned image has been obtained after printing and scanning an original digital copy, there exists a ground truth digital copy for each scanned version.

In this paper, we define a new inverse problem called \textbf{Descanning}, i.e., image restoration from a scanned copy to its original digital one.
Specifically, this refers to the restoration of information physically printed on papers that have been corrupted in the process of scanning or during preservation.
We broadly categorize degradation resulting from such processes into two types: color-related degradation (CD) and non-color-related degradation (NCD). 
CD contains color transition while NCD consists of external noise, internal noise, halftone pattern, texture distortion, and bleed-through effect, each of which will be explained in detail.

Although many real-world image restoration methods and datasets have been proposed, only a few have focused on various degradation mixtures that can exist in real-world scanned images due to the lack of scanned image datasets.
Therefore, it is crucial to acquire many real scanned images and examine their degradation characteristics systematically to train a learning-based descanning model.
In this study, we build a novel dataset for descanning, namely \textbf{DESCAN-18K}.
This is composed of 18,360 pairs of $1024\times 1024$ resolution RGB TIFF original images and scanned versions of them from various scanners.
DESCAN-18K provides rich information about the aforementioned six representative complex degradations in typical scanned images.
It also contains various natural scenes and texts, making the descanning task difficult yet practical.
These characteristics of our dataset differ from existing restoration datasets that usually have a single (or few) degradation type and contain \textit{either} texts or pictures.
We conduct a statistical analysis on DESCAN-18K as well as systematize the degradations existing within.
Based on this analysis, we also synthesize additional training data pairs to contain similar degradations as in the original DESCAN-18K.

Meanwhile, diffusion models \cite{firstdiff} have recently garnered attention as a highly effective generative method capable of performing low-level vision tasks \cite{ddrm, srdm}.
However, they are yet to be explored for restoring images with multiple degradations such as for our descanning problem.
To address such complex restoration problems, we propose a new image restoration model called \textbf{DescanDiffusion} consisting of the color encoder for global color correction and the conditional denoising diffusion probabilistic model (DDPM) \cite{ddpm} for local generative refinement.

Our main contributions can be summarized as follows: 

\begin{enumerate}

\item We define a novel practical image restoration problem, called descanning, which is to restore the original images by removing complex degradations present in the scanned images. 

\item We build DESCAN-18K, a large-scale dataset for the descanning task. 
We further conduct a statistical analysis of DESCAN-18K and analyze the degradation types resulting from various processes in converting original to scanned images.
Also, we devise a synthetic data generation scheme based on this analysis.

\item We propose DescanDiffusion, a new image restoration model composed of the color encoder and the conditional DDPM designed to address the descanning problem with multiple degradations.

\item We provide various experiments and analyses showing the effect of DescanDiffusion, including results on unseen-type scanners and comparison to commercial products. Our DescanDiffusion outperforms other baselines and generalizes well to new scenarios.

\end{enumerate}

\section{Related Works}

\subsection{Image Restoration with Single Degradation}

Most image restoration methods that handle single CD (e.g., color fading or saturation \cite{pix2pixHD, romnet, cyclegan, llflow}) have been developed based on the convolutional neural network (CNN) and Vision Transformer \cite{vit} \cite{uformer, restormer, swinir}.
For example, \cite{cyclegan} and \cite{pix2pixHD} are popular image-to-image translation generative adversarial network (GAN) \cite{gan} methods.
For single NCD, many image restoration methods have been proposed for a single task such as denoising \cite{denoising1, denoising2}, super-resolution (SR) \cite{sr1, sr2}, and deblurring \cite{deblurring1, deblurring2}.

These models show notable performance when only a single type (blur, noise, etc.) of degradation is present.
But it is unclear if they can handle many CDs and NCDs simultaneously.
In our descanning problem, scanned images have complex CDs and NCDs with high uncertainty and diversity due to digital processing stages, e.g., scanning, printing, etc.
Thus, directly restoring scanned images using the above methods may lead to poor performance, and a more dedicated model should be developed for descanning.
In this paper, we propose a novel image restoration model with components designed to adequately handle both CD and NCD.

\subsection{Real-world Photo Restoration} 
Many studies \cite{opr, dps, time, descreennet, uhdm, hdrunet} have been proposed for real-world photo restoration.
\cite{opr} uses translation networks for image and latent space, respectively, to restore real-world old photos with various degradations such as scratches, dust spots, and multiple noises.
\cite{dps} removes degradations from smartphone-scanned photos in a semi-supervised way, with smartphone-scanned DIV2K \cite{div2k} images as inputs and the original digital versions as targets.
\cite{uhdm} proposes ESDNet for demoiréing, which is a similar task to descanning in that both tasks aim to remove visually awkward color transitions and patterns simultaneously.

However, real-world scanned images still cannot be appropriately restored due to the more complex special NCDs such as the halftone pattern and bleed-through effect.
There are a few classic image processing-based methods for restoring scanned documents \cite{review, text}.
However, they mainly focus on eliminating dark borders and scanning shading which are dedicated to document-related degradations.
These degradations typically arise from the geometric misalignment of books (e.g., curled pages and book spines), which is different from our focus of comprehensively restoring scanned images containing a variety of color photos and texts to clean original (digital) images.

Hence, to holistically address the descanning problem, we build a huge dataset with real scanned images from multiple scanners and their originals. Also, we propose a descanning model that is tailored to the properties of scanned images.

\begin{figure*}
\centering
  \includegraphics[width=17.7cm]{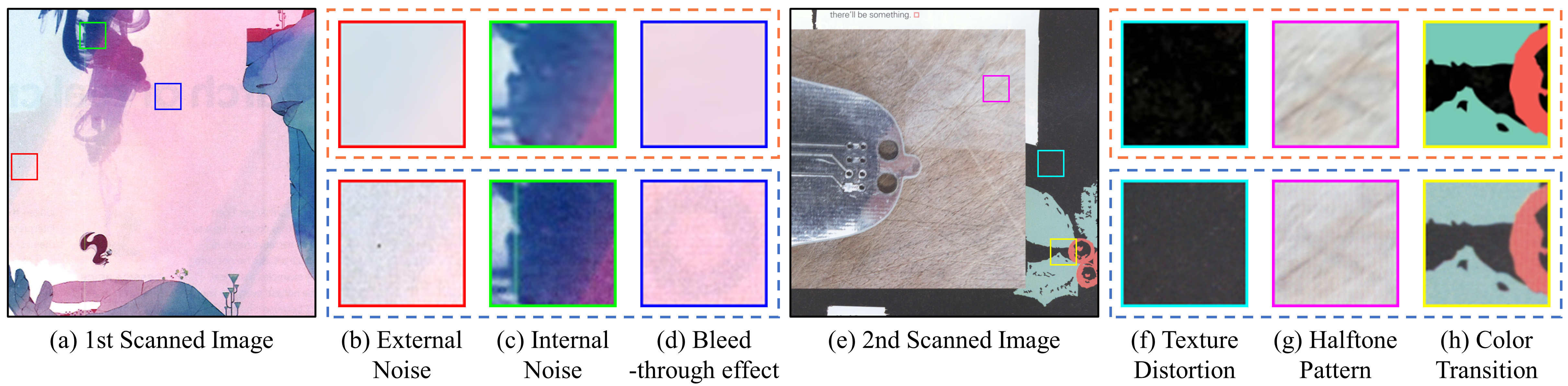}
  \caption{Examples of degradations in DESCAN-18K. Both (a) and (e) are scanned examples in DESCAN-18K. From (b) to (h), except for (e), patches in the upper row with orange dotted lines are from original images, and patches in the lower row with blue dotted lines are from their scanned counterpart (See the supplementary material for more diverse examples).}
  \label{fig:dataset_degradations}
\end{figure*}

\subsection{Diffusion Models for Image Restoration}

Recently, due to the impressive generation performance of diffusion models, they have been actively applied to various fields such as text-to-image generation \cite{dalle, photorealistic}, natural language processing \cite{diffnlp}, and vision applications \cite{repaint,imgseg}.
Several diffusion models have also been developed for image restoration. \cite{ddrm} introduces a diffusion model for various image restoration tasks such as SR, deblurring, and inpainting. \cite{srdm} adapts DDPM in a conditional manner and achieves strong SR performance with iterative refinement processes.

In this paper, we propose DescanDiffusion which exploits the restoration power and generalization ability of diffusion models, especially DDPM. 
We observed that naively applying vanilla DDPMs for descanning can result in shifting away from the color distribution of the original image.
To tackle this issue, we design a color encoder that predicts the color distribution of the original image given the scanned image and computes the color-corrected image, thereby offering a superior starting point for DDPM. The estimated color distribution is also used as a condition for the diffusion model to explicitly guide the model with color information during the diffusion process.

\section{Dataset}
\label{sec:Dataset}

In this work, we introduce a large-scale dataset named DESCAN-18K that contains 18,360 pairs of scanned and original images of $1024\times 1024$ resolution in an RGB TIFF format. In order to acquire a large amount of scanned and original image pairs, we use the 11 types of magazines from Raspberry Pi Foundation \cite{raspberry} licensed under CC BY-NC-SA 3.0, which contain diverse image/text contents, colors, textures, etc.
They also include various types of degradations due to the sufficiently long preservation duration, i.e., from a few days to seven years.

\subsection{Dataset Processing} 

We manually scanned each page of the magazines with different popular scanners: Plustek OpticBook 4800, Canon imageRUNNER ADVANCE 6265, Fuji Xerox ApeosPort C2060, and Canon imagepress C650.
The scanned images are digitized in the format of RGB TIFF and calibrated by the IT 8.7 (ISO 12641) standard. Since most scanners follow this standard for color calibration, it reduces the variance across scanner models, making our model more generalizable to different scanner types.
After obtaining scanned images, we gather their corresponding original PDF copies online and convert them to the same RGB TIFF format.

As the scanned and original versions of magazine pages are misaligned due to margin settings and crumpled pages, etc., we take the following steps to align them:
we first perform image registration with AKAZE \cite{akaze} for each page.
The page pairs are then manually inspected, filtering out images that are unmatched on a significant scale.
Finally, we randomly crop each image into $1024\times 1024$ sizes and register them again with AKAZE, securing 18,360 pairs of aligned scanned and original images.

Among 18,000 images scanned using Plustek OpticBook 4800 and Cannon imageRUNNER ADVANCE 6265, 17,640 are used for training and 360 are used for validation.
That is, the validation set is different from the training set.
We leave 360 images scanned by Fuji Xerox ApeosPort C2060 and Canon imagepress C650 as the testing set.
Note that the scanners used for the testing set are \textit{different} from those for training and validation, which allows us to evaluate the generalization ability for unseen-type scanners.

\subsection{Dataset Analysis} 
\label{sec:Dataset Analysis}

By analyzing the complete dataset, we classify the degradations in scanned images into six types. 
Note that although we discuss each type of degradation separately, degradations themselves are often a combination of multiple degradation types. In Fig. \ref{fig:dataset_degradations}, both (a) and (e) are scanned examples of DESCAN-18K. 
From Fig. \ref{fig:dataset_degradations} (b) to (h), except for (e), patches in the upper row with orange dotted lines are from the original images, and patches in the lower row with blue dotted lines are from their scanned counterparts. 

As Fig. \ref{fig:dataset_degradations} shows, we categorize degradations as follows:

\begin{itemize}

  \item \textbf{External noise} is caused by the inflow of foreign substances during printing, scanning, and preserving.
It appears in the form of dots or localized stains.

  \item \textbf{Internal noise} is the visual degradation generated by the scanning process. It usually occurs as crumpled, curved and/or linear laser patterns.

  \item \textbf{Bleed-through effect} is a degradation in which the contents of the back page are transmitted through and scanned together.
Note that it solely appears in scanned images, not in ordinary real-world images.

  \item \textbf{Texture distortion} consists of physical textures or wrinkles that occur during scanning.
Note that this tends to appear globally, whereas external noise tends to appear locally in a specific region.

  \item \textbf{Halftone pattern} is generated due to the printing process where many dots of different colors (e.g., cyan, magenta, yellow and black), sizes and spacings are imprinted to represent continuous shapes. 

  \item \textbf{Color transition} is the chromatic distortion of an image being globally altered during scanning and preserving.
There are degradations such as color fading or saturation.

\end{itemize}

Detailed statistical analysis on DESCAN-18K can be found in the supplementary material. 

\begin{figure*}[hbt!]
\centering
  \includegraphics[width=16.5cm]{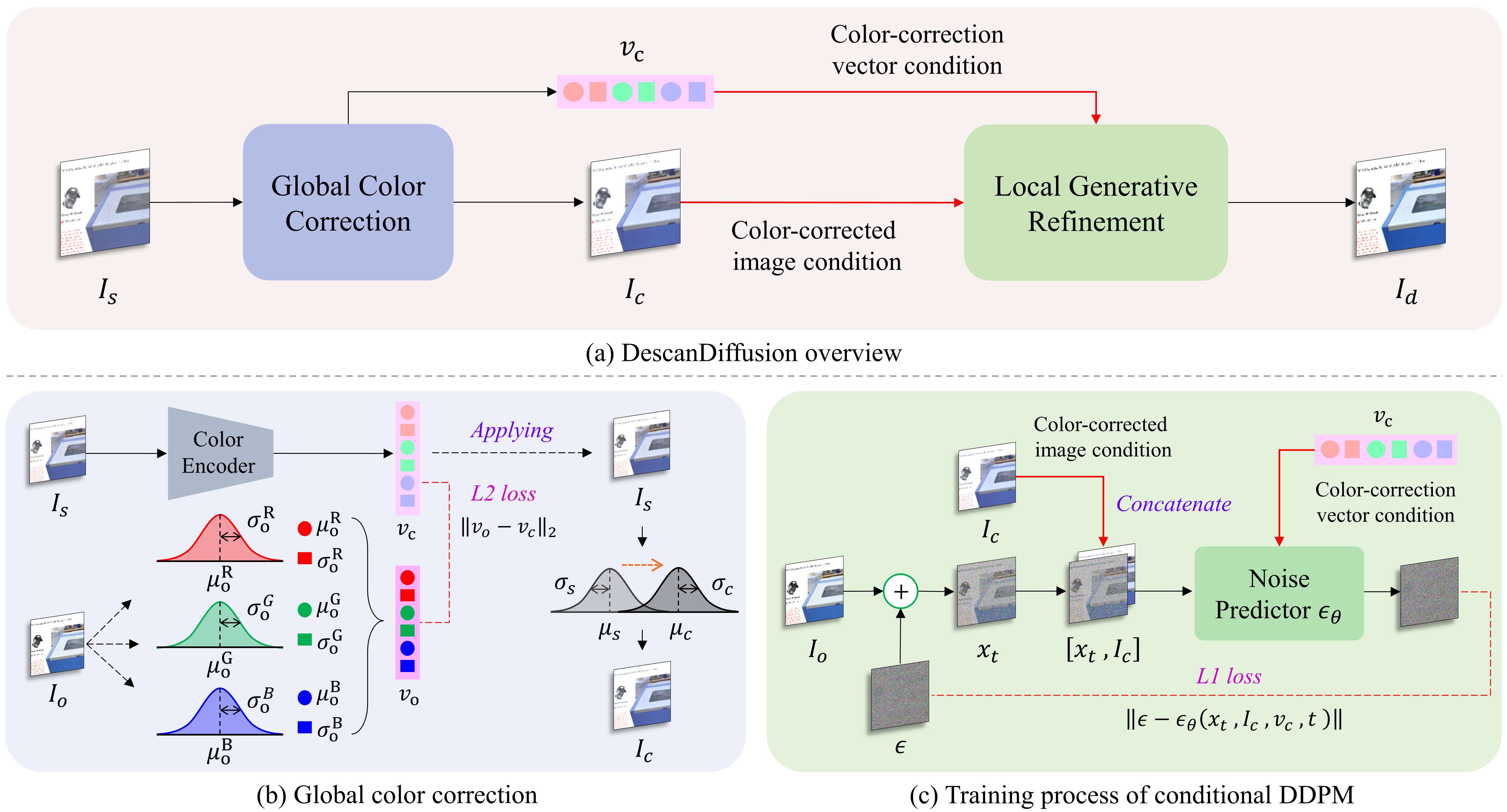} 
  \caption{Overview of our DescanDiffusion: (a) the whole process of DescanDiffusion with global color correction and local generative refinement modules; (b) global color correction module with a color encoder that predicts the color correction vector $v_c$ and produces the color-corrected image $I_c$; (c) the training process of the local generative refinement module with a conditional DDPM.} 
  \label{fig:overview}
\end{figure*}

\subsection{Synthetic Data Generation}

Based on our analysis of the dataset, we simulate some of the degradations found in scanned images: (i) for color transition, we modify the HSV color space of the original image, (ii) for the bleed-through effect, we alpha-blend two original images, (iii) for halftone pattern and texture distortion, we apply Gaussian noise, (iv) for external and internal noise, we synthesize the form of dots and linear laser patterns, respectively.
By doing so, we aim to improve the generalization performance of DescanDiffusion, enabling it to effectively restore images even if they are scanned from new scanners.

The degradation strength and probability of synthesizing them are determined randomly for each sample, with a uniform distribution, and original images from a subset of the DESCAN-18K training dataset are utilized to generate synthetic data. 
Specifically, we train DescanDiffusion+ using 25\% synthetic-original pairs and 75\% scanned-original pairs out of the total 17,640 image pairs in the training set, while the original DescanDiffusion exclusively utilizes scanned-original pairs from the same set.
This ratio was determined empirically and its ablation study is provided in the supplementary material.
Note that our synthetic data generation scheme can be applied to any original document image to augment the training set further.

\section{Preliminary: DDPM}

In this section, we briefly introduce DDPM \cite{ddpm}, an important element of DescanDiffusion. Given an image $x_o$ from a data distribution, a forward noising diffusion Markov process is applied, adding the noise gradually in multiple steps $t$, where the level of noise is controlled by a noise schedule \(\beta\), yielding

\begin{equation}
q\left(x_{1: T} \mid x_0\right)=\prod_{t=1}^T q\left(x_t \mid x_{t-1}\right)
\end{equation}
\begin{equation}
q\left(x_t \mid x_{t-1}\right)=N\left(x_t ; \sqrt{1-\beta_t} x_{t-1}, \beta_t I\right)
\end{equation}
where $T$ is the total number of steps in the diffusion process. \(x_0\) is a sample from the data distribution and the \(x_0, x_1, \ldots ,x_T\) are the latent variables. 
As $T \rightarrow \infty$, $x_T$ converges to a Gaussian isotropic noise.
Any latent space $x_t$ can be sampled during the forward process, using the following closed-form formulation where $t$ is drawn from
$\forall t \sim \mathcal{U}(\{1, \ldots, T\})$:
\begin{equation}
\label{eqn_diff:04}
x_t=\sqrt{\bar{\alpha}_t} x_0+\epsilon \sqrt{1-\bar{\alpha}_t} \\
\end{equation}
where $\epsilon \sim \mathcal{N}(0, I)$, $\alpha_t=1-\beta_t$ and $\bar{\alpha}_t=\prod_{i=1}^t \alpha_i$.

In order to generate a clean output, a reverse denoising diffusion process of estimating $q(x_{t-1} \mid x_t)$ is performed. We learn the reverse process $p_\theta$ utilizing a neural network parameterized by $\theta$ as 
\begin{equation} 
p_\theta\left(x_{t-1} \mid x_t\right) = \mathcal{N}\left(x_{t-1} ; \mu_\theta\left(x_t, t\right),{ \sigma_\theta\left(x_t, t\right)}^2I\right)
\end{equation}
where $\mu_\theta\left(x_t, t\right)$ is the estimated mean, and $\sigma_\theta\left(x_t, t\right)^2$ is the estimated variance that can be fixed as \(\beta_t\). In DDPM, instead of training \(\mu_\theta\), a neural network \(\epsilon_\theta\) is trained to estimate $\epsilon$  given $x_{t}$.
The \(\epsilon_\theta\) is trained by minimizing the following loss:
\begin{equation}
L_{e r r}=\mathbb{E}_{x_o, t, \epsilon \sim \mathcal{N}(0, I)}\left[\left\|\epsilon-\epsilon_\theta\left(x_t, t\right)\right\|\right]
\end{equation}

In general, for inference, we start with sampling \(x_T \sim \mathcal{N}(0, I)\), and then iteratively refine the latent variable $x_t$ to generate $x_{t-1}$, ultimately obtaining $x_o$ at $t=0$.

\section{Proposed Method}
\label{sec:Methods}

Since complex degradations are mixed in the scanned image, descanning is more challenging than other image restoration tasks.
As we categorize the degradations in scanned images into CD and NCD, we design a new image restoration model DescanDiffusion that consists of two modules: (i) a global color correction module; and (ii) a local generative refinement module, which deals with CD and NCD, respectively. Fig. \ref{fig:overview} shows an overview of our proposed DescanDiffusion.

\subsection{Global Color Correction with the Color Encoder}

In the global color correction module shown in Fig. \ref{fig:overview} (b), we utilize the color encoder $\Phi$ to predict the color distribution of the original image $I_o$. The output of $\Phi$ is then used to correct the color distribution of the scanned image $I_s$ such that the color distribution of $I_s$ is approximated to that of $I_o$ thus removing most CDs from $I_s$.
This results in the color-corrected image $I_c$, which can be exploited as a good condition in the following local generative refinement module.

We adopt ResNet-34 \cite{resnet} as $\Phi$ because it is computationally efficient while having a large receptive field.
With $I_s$ as input and the color distribution of the original image ($v_o \in \mathbb{R}^{1\times6}$) as the target, $\Phi$ predicts $v_c=\Phi(I_s)$, where $v_c \in \mathbb{R}^{1\times6}$.
Here, $v_o$ and $v_c$ are vectors composed of means ($\mu^k_o$, $\mu^k_s$) and standard deviations ($\sigma^k_o$, $\sigma^k_s$) of color channels $k$ in $I_o$ and $I_c$, respectively, where $k \in\{R, G, B\}$.
This process is optimized by the L2 loss which can be written as
\begin{equation}
\label{eqn:04}
    {L}_{2}(\Theta) = \|v_o-v_c\|_2,
\end{equation}
where $\Theta$ denotes the learnable parameters of $\Phi$.

Employing the estimated color statistics, i.e., $\mu^k_c$ and $\sigma^k_c$, we re-normalize the color distribution of $I_s$ to mimic the color distribution of $I_o$.
This re-normalization process can be formulated as
\begin{equation}
\label{eqn:09}
\begin{split}
I^k_c =  \frac{I^k_s - \mu^k_s}{\sigma^k_s + \varepsilon }  \sigma^k_c + \mu^k_c,
\end{split}
\end{equation}
where $I^k_c$ and $I^k_s$ are the $k$-th channels in $I_c$ and $I_s$, respectively. $\varepsilon $ in Eq. \ref{eqn:09} is to secure numerical stability when $\sigma^k_s$ is close to zero, and set to $2^{-16}$.
We perform this re-normalization for each R, G, B channel and concatenate them to be $I_c$. 

It is noted that image-to-image translation methods \cite{pix2pix, cyclegan} that are able to mimic histogram matching can also be used to restore $I_c$. However, we found that the proposed color correction method yields competitive performance with much lower computational complexity.

\subsection{Local Generative Refinement with DDPM} 

Our proposed Local Generative Refinement Diffusion Model (LGRDM) mainly aims at removing NCDs from the color-corrected image $I_c$. In addition, LGRDM allows shifting the local color distributions of $I_c$ further toward $I_o$.

\SetKwInput{KwInput}{Input}                
\SetKwInput{KwInitialize}{Initialize}              
\SetKwInput{KwRepeat}{Repeat}              
\SetKwInput{KwUntilconverged}{Until Converged}   
\SetKwInput{kwLoad}{Load}

\begin{algorithm}[H]
\DontPrintSemicolon
\label{Algo:Train}
  \KwInput{Scanned images and corresponding original images pairs, $P = \{(I_s^n, I_o^n)\}_{n=1}^N$ and total number of diffusion steps, $T$}
  \KwInitialize{Pre-trained color encoder $\Phi$ and randomly initialized conditional denoising network $\epsilon_\theta$}
  \KwRepeat{  
  
  \text{1: } Sample scanned and original image pairs \((I_s , I_o) \sim P\)
  
  \text{2: } $v_c = \Phi(I_s)$
  
  \text{3: } $I_c = \text{ReNormalize}(v_c, I_s)$ in Eq. \ref{eqn:09}
  
  \text{4: } Sample $\epsilon \sim \mathcal{N}(0, I)$, $ t \sim \mathcal{U}(\{1, \ldots, T\})$
  
  \text{5: } Take gradient step on:\\           
     $\nabla_\theta\|\epsilon-\epsilon_\theta(x_t, I_c, v_c, t)\|$, $x_t=\sqrt{\bar{\alpha}_t} I_o +\epsilon\sqrt{1-\bar{\alpha}_t}$
     }
\textbf{Until Converged}
\caption{Training of LGRDM}\label{Algo:Train}
\end{algorithm}

\begin{algorithm}[H]
\DontPrintSemicolon
  
\KwInput{Scanned image $I_s$ and the optimal number of sampling steps $T_o$, where $T_o \le T$}
\kwLoad{Pre-trained color encoder $\Phi$ and conditional denoising network $\epsilon_\theta$}

\text{1: } $v_c = \Phi(I_s)$

\text{2: } $I_c = \text{ReNormalize}(v_c, I_s)$ in Eq. \ref{eqn:09}

\text{3: } $x_{T_o} = I_c$
  
  \For{$t = T_o, T_{o-1}, \dots , 1$}
  {
  If $t > 1$ then Sample $z \sim \mathcal{N}(0, I)$ else $z = 0$\\
  $x_{t-1} = \frac{1}{\sqrt{\alpha_t}} \left(x_t - \frac{1-\alpha_t}{\sqrt{1-\bar{\alpha}_t}} \epsilon_\theta(x_t,I_c,v_c, t)\right) + \sigma_t z$
  }
\caption{Inference of LGRDM}
\label{Algo:Inference}
\end{algorithm}

LGRDM involves a conditional denoising network $\epsilon_\theta$ based on UNet \cite{unet}. As shown in Fig. \ref{fig:overview} (c), $\epsilon_\theta$ is conditioned on two factors from the previous global color correction module: the color-corrected image $I_c$, and the color correction vector $v_c$. 

The first condition $I_c$ guides the restoration process toward $I_o$ resulting in faster and better convergence. For $I_c$ conditioning, we concatenate $I_c$ with the latent variables $x_t$ at each time step $t$, where $t \in \{T, \ldots ,1\}$.

The second condition $v_c$ aims to constrain color distribution shifts of the generated image. Note that DDPM tends to generate different color distributions from the target image due to its high generation ability. Color conditioning with $v_c$ serves as color guidance, allowing to preserve consistent color distribution. For conditioning with $v_c$, we project $v_c$ to a higher dimensional embedding space with a single-layer color projection network. The resulting color embedding is then added to the timestep embedding for conditioning. \cite{nichol2021improved} 

Finally, $\epsilon_\theta$ is trained to estimate the added noise in $x_t$, where $x_t=\sqrt{\bar{\alpha}_t} I_o +\epsilon\sqrt{1-\bar{\alpha}_t}$. 
This process is optimized with the following loss: 
\begin{equation}
L_{e r r}=\mathbb{E}_{x_0, t, \epsilon \sim \mathcal{N}(0, I),I_c, v_c}\left[\left\|\epsilon-\epsilon_\theta\left(x_t, t, I_c, v_c\right)\right\|\right]
\end{equation}

Algorithm \ref{Algo:Train} and \ref{Algo:Inference} describe the pseudo-codes of the training and inference processes of LGRDM, respectively. 
Note that $T_o$ in Algorithm \ref{Algo:Inference} is the optimal number of sampling steps and is determined empirically. (Detailed explanation can be found in the supplementary material)

\begin{table}[ht]
\setlength{\tabcolsep}{4pt}
\centering\fontsize{9}{11}\selectfont  
  \begin{tabular}{lcccc}
    \hline
    Method & PSNR (dB) $\uparrow$ & SSIM $\uparrow$ & LPIPS $\downarrow$ & FID $\downarrow$ \\
    \hline
    Pix2PixHD & 20.58 & 0.8014 & 0.057 & 18.30 \\
    CycleGAN & 21.52 & 0.8417 & 0.050 & 16.99 \\
    HDRUNet & 20.90 & 0.8480 & 0.055 & 16.42 \\
    Restormer & 20.37 & 0.7915 & 0.152 & 25.57 \\
    ESDNet & 21.22 & 0.8418 & 0.088 & 15.24 \\
    NAFNet & 22.03 & 0.8538 & 0.048 & 16.00 \\
    OPR* & 18.09 & 0.7249 & 0.158 & 21.45 \\
    DPS* & 17.93 & 0.7354 & 0.150 & 41.64 \\    
    \hline
    Clear Scan & 21.46 & 0.8183 & 0.054 & 18.09 \\
    Adobe Scan & 15.80 & 0.6153 & 0.141 & 23.55 \\
    Microsoft Lens & 20.48 & 0.8013 & 0.056 & 18.97 \\
    \hline
    DescanDiffusion & {23.40} & {0.8717} & \textbf{0.042} & \textbf{13.51} \\
    DescanDiffusion+ & \textbf{23.43} & \textbf{0.8736} & {0.044} & {14.60} \\
    \hline
  \end{tabular}
  \caption{Quantitative comparison of descanning performance on original DESCAN-18K testing set (average PSNR/SSIM/LPIPS/FID). Methods with an asterisk(*) are pre-trained versions.}
  \label{tab:quantitative_comparison}
\end{table}

\subsection{Discussion on the Training Strategy}

We trained our model from scratch with our DESCAN-18K. The ResNet is trained separately so that it can serve the purpose of global color correction by aligning the color distribution of the scanned image with the original one. If our full framework is trained jointly from the start, premature outputs of the ResNet may confuse the training of the DDPM leading to sub-optimal results, since the DDPM is conditioned on outputs of the ResNet. It is similar to the common training strategy of freezing the text encoder during training text-to-image diffusion models \cite{imagen}.

\begin{figure*}[hbt!]
\centering
  \includegraphics[width=16.58cm]{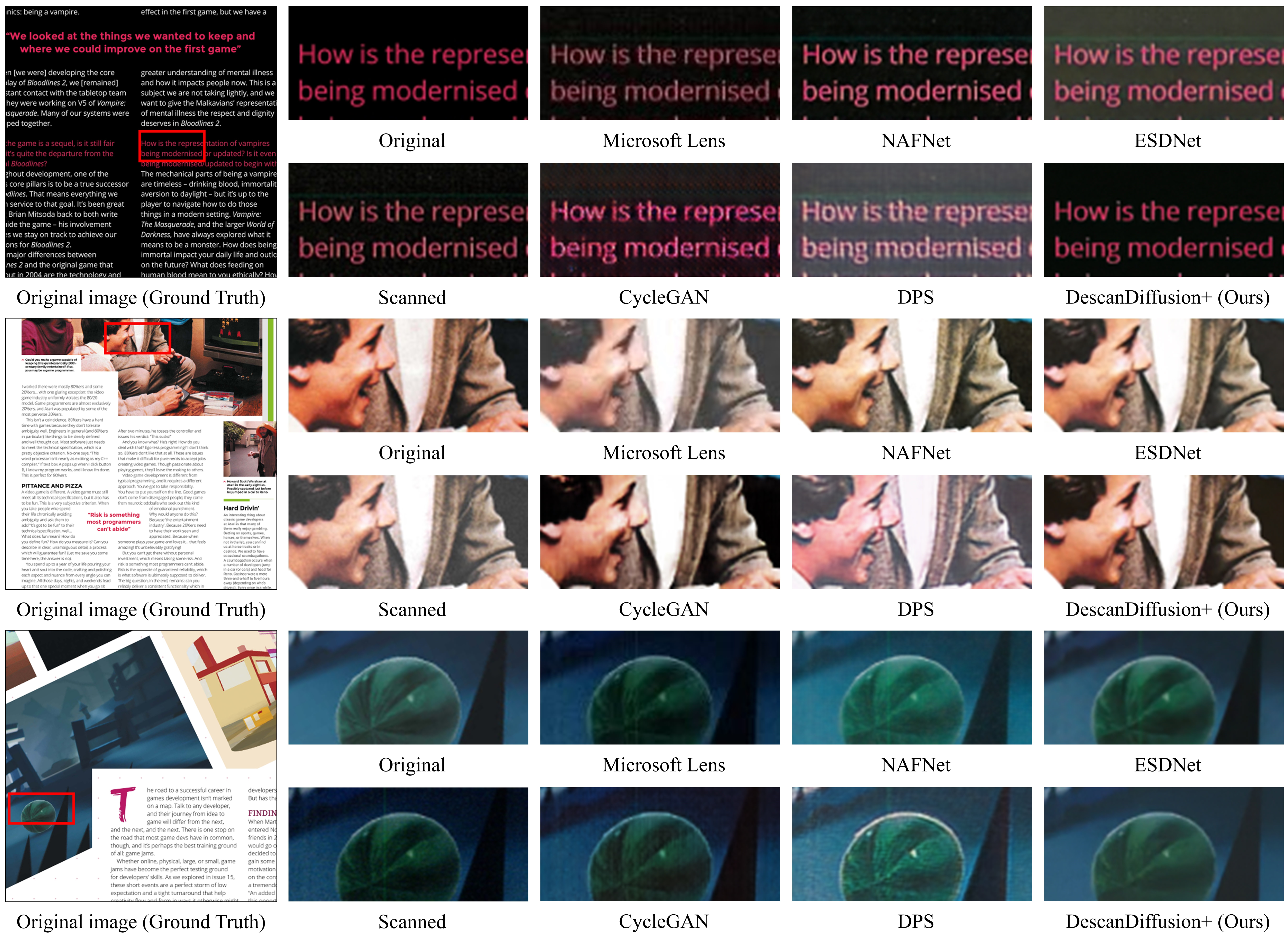}
  \caption{Qualitative comparisons of descanning performance on DESCAN-18K testing set. Scanned images (denoted as Scanned) in each row mostly have the following degradations; $1^{st}$ row: texture distortion, color transition, and internal noises in a linear laser form, $2^{nd}$ row: color transition and texture distortion, $3^{rd}$ row: same degradations in the $1^{st}$ row. Our DescanDiffusion+ model outperforms another image-to-image translation model, real-world photo restoration model, recent image restoration model, and commercial product in handling degradations in text regions, natural scenes, and screen contents (See the supplementary material for more diverse examples).}
  \label{fig:comparison}
\end{figure*}

\section{Experiments}
\label{sec:Experiments}

\subsection{Experimental Setup} 

Given that descanning is a novel problem that has yet been previously explored, comparing to existing work is highly challenging. To address this problem, we extensively evaluate our method with models performing related tasks, which can be classified into: (i) image-to-image translation models (Pix2PixHD \cite{pix2pixHD} and CycleGAN \cite{cyclegan}), (ii) recent image restoration models that conduct similar tasks as descanning (HDRUNet \cite{hdrunet}, Restormer \cite{restormer}, ESDNet \cite{uhdm}, and NAFNet \cite{NAFNet}), (iii) real-world photo restoration models (OPR \cite{opr} and DPS \cite{dps}), (iv) commercial products (Clear Scan\footnote{Accessed 20 July. 2023, \url{https://play.google.com/store/apps/details?id=com.indymobileapp.document.scanner}} \cite{cleanscan}, Adobe Scan\footnote{Accessed 20 July. 2023, \url{https://play.google.com/store/apps/details?id=com.adobe.scan.android}} \cite{abobescan}, and Microsoft Lens\footnote{Accessed 20 July. 2023, \url{https://play.google.com/store/apps/details?id=com.microsoft.office.officelens}} \cite{microsoftlens}), (v) a recent diffusion-based image restoration model (DDRM \cite{ddrm}).

\begin{table}[ht]
\setlength{\tabcolsep}{4pt}
\centering\fontsize{9}{11}\selectfont
  \begin{tabular}{lcccc}
    \hline
    Method & PSNR (dB) $\uparrow$ & SSIM $\uparrow$ & LPIPS $\downarrow$ & FID $\downarrow$ \\
    \hline
    Pix2PixHD & 21.49 & 0.8378 & 0.050 & 18.18 \\
    CycleGAN & 23.16 & 0.8640 & 0.043 & 17.57 \\
    HDRUNet & 22.49 & 0.8627 & 0.048 & 19.55 \\
    Restormer & 21.96 & 0.8238 & 0.088 & 26.80 \\
    ESDNet & 22.05 & 0.8686 & 0.059 & 16.27 \\
    NAFNet & 23.14 & 0.8714 & \textbf{0.040} & 16.44 \\
    OPR* & 19.47 & 0.7385 & 0.188 & 29.70 \\
    DPS* & 19.60 & 0.7724 & 0.100 & 36.29 \\ 
    \hline
    Clear Scan & 23.16 & 0.8450 & 0.047 & 18.39 \\
    Adobe Scan & 16.41 & 0.6379 & 0.127 & 21.75 \\
    Microsoft Lens & 21.46 & 0.8202 & 0.053 & 19.04 \\
    \hline
    DescanDiffusion & {23.40} & {0.8717} & {0.042} & \textbf{13.51} \\ 
    DescanDiffusion+ & \textbf{23.43} & \textbf{0.8736} & {0.044} & {14.60} \\ 
    \hline
  \end{tabular}
  \caption{Quantitative comparison of descanning performance on DESCAN-18K testing set with global color correction by histogram matching. We used the same metrics and models as in Table.~\ref{tab:quantitative_comparison}.}
  \label{tab:quantitative_comparison_histo}
\end{table}

\begin{table}[ht]
\setlength{\tabcolsep}{4pt}
\centering\fontsize{9}{11}\selectfont
  \begin{tabular}{lcccc}
    \hline
    Method & PSNR (dB) $\uparrow$ & SSIM $\uparrow$ & LPIPS $\downarrow$ & FID $\downarrow$ \\
    \hline
    DDRM & 24.14 & 0.8800 & 0.034 & 17.78 \\
    \hline
    DescanDiffusion & \textbf{25.72} & \textbf{0.9155} & \textbf{0.026} & \textbf{13.83} \\
    \hline
  \end{tabular}
  \caption{Quantitative comparison of descanning performance on original DESCAN-18K testing set between DDRM and our DescanDiffusion. Because the DDRM works only at a resolution of $256\times256$, performance comparisons for DDRM are conducted exclusively at this resolution. We used the same metrics as in Table.~\ref{tab:quantitative_comparison}.}
  \label{tab:ddrm}
\end{table}

We re-train all compared methods using DESCAN-18K training set except OPR and DPS, for which official pre-trained models are used as we expect that they are optimized for restoring damaged real-world photos.

\subsection{Comparison to Existing Methods}

We employ the following four metrics to quantitatively evaluate the descanning performance.
PSNR is adopted to calculate pixel-wise fidelity between the restored and original image.
To measure perceptual quality, we use SSIM \cite{ssim} and LPIPS \cite{lpips}. 
We also calculate Fréchet Inception Distance (FID) \cite{fid} to assess generation performance.

Quantitative results are reported in Table.~\ref{tab:quantitative_comparison}. Our DescanDiffusion and DescanDiffusion+ outperform other methods including commercial products on all metrics.
As the testing set \textit{only} contains images scanned from different scanners that were \textit{not} used in the training set, the results suggest that our proposed method has good generalization performance and practicality for unseen-type scanners.
In other words, regardless of which scanner is used, our method is able to restore scanned images robustly, which is important for the descanning task as various scanners exist in the real-world.

Table.~\ref{tab:quantitative_comparison_histo} shows quantitative results after applying the global color correction through histogram matching on compared models.
Compared to Table.~\ref{tab:quantitative_comparison}, most models show notable improvements in most metrics after applying a global color correction.
This suggests that CDs are dominant in scanned images, emphasizing the importance of addressing them with global color correction.

It also can be interpreted that the proposed color encoder and the color-conditioned DDPM contribute to the high descanning performance by estimating low dimensional color statistics and guiding the model with the color distribution.

Compared to DescanDiffusion, DescanDiffusion+ provides slightly better performance in PSNR and SSIM. For LPIPS and FID, DescanDiffusion and DescanDiffusion+ result in comparable performance. We found that DescanDiffusion+ tends to better eliminate high-frequency degradations which are similar to the synthesized ones when analyzed visually (See the supplementary material).

In addition, Table.~\ref{tab:ddrm} demonstrates that our DescanDiffusion surpasses the DDRM \cite{ddrm}, a recent diffusion-based image restoration model. This observation implies that diffusion-based image restoration models necessitate additional functions, such as our proposed global color correction module, to effectively eliminate multiple degradations in scanned images.

Fig. \ref{fig:comparison} shows visual results of both deep-learning-based methods and commercial products.
DescanDiffusion almost resolves NCD and CD problems in scanned images, while the others leave these issues inadequately resolved or even worsen them.
For instance, in the example in $3^{rd}$ row, NAFNet and ESDNet cannot completely eliminate internal noises.
Moreover, the commercial product and real-world photo restoration models are not able to remove degradation well or even generate additional artifacts in some cases.

\begin{table}[ht]
\centering\fontsize{9}{11}\selectfont
  \begin{tabular}{lccccc}
    \hline
     & (a) & (b) & (c) & (d) & (e) \\
    \hline
    CIC & \xmark & \cmark & \cmark & \cmark & \xmark  \\
    CVC & \xmark & \xmark & \cmark & \cmark & \xmark  \\
    SDG & \xmark & \xmark & \xmark & \cmark & \cmark  \\ 
    \hline
    PSNR (dB) & 22.72 & 23.18 & 23.40 & \textbf{23.43} & 23.09  \\ 
    SSIM & 0.8583 & 0.8652 & 0.8717 & \textbf{0.8736} & 0.8672  \\ 
    \hline
  \end{tabular}
  \caption{Ablation study of three components in the proposed method.}
  \label{tab:abl3}
\end{table}

\begin{table}[ht]
\centering\fontsize{9}{11}\selectfont
  \begin{tabular}{l|c}
    \hline
    Method & Inference Time  \\
    \hline
    CycleGAN & $10^{-5}$s  \\
    Restormer & 0.5289s  \\
    ESDNet & 0.2251s  \\
    NAFNet &  0.0013s  \\
    \hline
     DescanDiffusion & 2.5827s  \\
    \hline
  \end{tabular}
  \caption{The inference time comparison on DESCAN-18K testing set.}
  \label{tab:infertime}
\end{table}

\subsection{Ablation Study}
\label{subsec:abl}

We conduct an ablation study to analyze the effect of three components in our proposed model: (i) color-corrected image condition for DDPM (denoted as CIC), (ii) color-correction vector condition for DDPM (denoted as CVC), and (iii) synthetic data generation scheme (denoted as SDG).

Table. \ref{tab:abl3} (a) and (b) shows that using the color-corrected image ($I_c$) obtained through the global color correction module as a condition for DDPM leads to a significant performance boost compared to the vanilla DDPM conditioned by the scanned image.
Additionally providing the color-correction vector ($v_c$) from the global color correction module as a condition to DDPM further improves the descanning performance. (Table. \ref{tab:abl3} (c))
The color-correction vector is composed of the mean and standard deviation from each R, G, B channel in the color-corrected image.
Hence, it can explicitly guide DDPM to consistently maintain the color distribution of the color-corrected image.
Finally, we mix synthetic data rather than using only the original DESCAN-18K to enhance the generalization ability of DescanDiffusion when handling input images scanned by unseen-type scanners.
Table. \ref{tab:abl3} (d) demonstrates that our model with SDG shows superior performance compared to the other versions.
Meanwhile, when applying only SDG to the vanilla DDPM (Table. \ref{tab:abl3} (e)), it outperformed the vanilla DDPM. However, compared to our final model (Table. \ref{tab:abl3} (d)) including CIC, CVC, and SDG, a performance drop is observed. Therefore, it is verified that global color correction is still important.

\subsection{Inference Time Evaluation}

Table. \ref{tab:infertime} provides a comparison of inference times for representative competitor models, conducted on an NVIDIA TESLA V100 GPU. As our method is a diffusion-based model, the inference time is slower compared to others based on CNNs or Transformers. However, as illustrated in Table. \ref{tab:quantitative_comparison}, our model exhibits superior performance compared to other methods. Moreover, in our training process, sampling begins from the scanned image instead of pure noise. Consequently, our method's inference time can be reduced by 92\% with just 10 reverse steps. (Detailed explanation can be found in the Algorithm \ref{Algo:Inference} and the supplementary material)

\subsection{Experiment on Additional Datasets}
\begin{table}[ht]
\centering\fontsize{9}{11}\selectfont  
  \begin{tabular}{lccc}
    \hline
    Method & NRQM $\uparrow$ & NIQE $\downarrow$ & PI $\downarrow$  \\
    \hline
    CycleGAN & 6.35/6.56 & \textbf{4.03}/3.68 & \textbf{3.88}/3.57  \\ 
    Restormer & \textbf{6.62}/6.43 & 8.92/7.56 & 6.17/5.58  \\
    ESDNet & 5.72/6.42 & 4.50/3.64 & 4.35/3.68  \\
    NAFNet & 4.86/6.47 & 4.98/4.02 & 5.06/3.80 \\
    \hline
     DescanDiffusion & 6.30/\textbf{6.72} & 4.77/\textbf{3.40} & 4.22/\textbf{3.37}  \\
    \hline
  \end{tabular}
  \caption{Quantitative comparison of descanning performance on DPS and OPR datasets (DPS/OPR). Since these datasets lack clear reference images, we utilized non-reference image quality metrics (average NRQM \cite{nrqm} / NIQE \cite{niqe} / PI \cite{pi}).
  }
  \label{tab:adddata}
\end{table}

To evaluate the performance of our proposed method on various image degradations, we further compared our model on additional datasets: 100 smartphone-scanned images from DPS \cite{dps} and 7 old photo images from OPR \cite{opr}. Table. \ref{tab:adddata} shows the quantitative results of each dataset (separated by a slash) for comparison models, validating that our DescanDiffusion generalizes well to smartphone-scanned images and old photos containing multiple degradations. Nevertheless, the specialization of our DescanDiffusion lies in removing mixtures of complex NCDs and CDs unique to scanned images from scanners. Furthermore, due to such characteristics of scanned images, it is noted that our DESCAN-18K is the most suitable dataset for evaluating the descanning performance.

\section{Conclusion}
Restoring scanned images is crucial in the digital world due to the vast amount of scanned content. To the best of our knowledge, we are the first to define this problem as descanning. In order to address this problem, we introduce a new large-scale dataset called DESCAN-18K that includes pairs of scanned and original images. Additionally, we classify the degradation types in DESCAN-18K into two categories: CD and NCD. Based on the analysis of degradation types, we propose a new image restoration model called DescanDiffusion, which utilizes a combination of the color encoder for global color correction and the conditional DDPM for local generative refinement. Thanks to the informative dataset and a dedicated model, DescanDiffusion achieves remarkable performance in terms of the visual quality of restored images. We believe that our work paves the way to handle restoration problems having highly complex and various degradations by offering detailed analyses and effective architecture design strategies.
Lastly, applying our proposed model to enhance downstream tasks like optical character recognition (OCR) or extending the application of our proposed dataset to evaluate new real-world image restoration models can be important future directions.

\section{Acknowledgments}

This work was supported in part by the Institute of Information and Communications Technology Planning and Evaluation (IITP) Grant funded by the Korea Government (MSIT) under Grant 2022-0-00759, in part by the Institute of Information and Communications Technology Planning and Evaluation (IITP) grant funded by the Korea Government (MSIT) (Artificial Intelligence Innovation Hub) under Grant 2021-0-02068, and in part by the Institute of Information \& communications Technology Planning \& Evaluation (IITP) grant funded by the Korea government (MSIT) (No.RS-2022-00155911, Artificial Intelligence Convergence Innovation Human Resources Development (Kyung Hee University)).

\bibliography{aaai24}

\clearpage

\includepdf[pages=-]{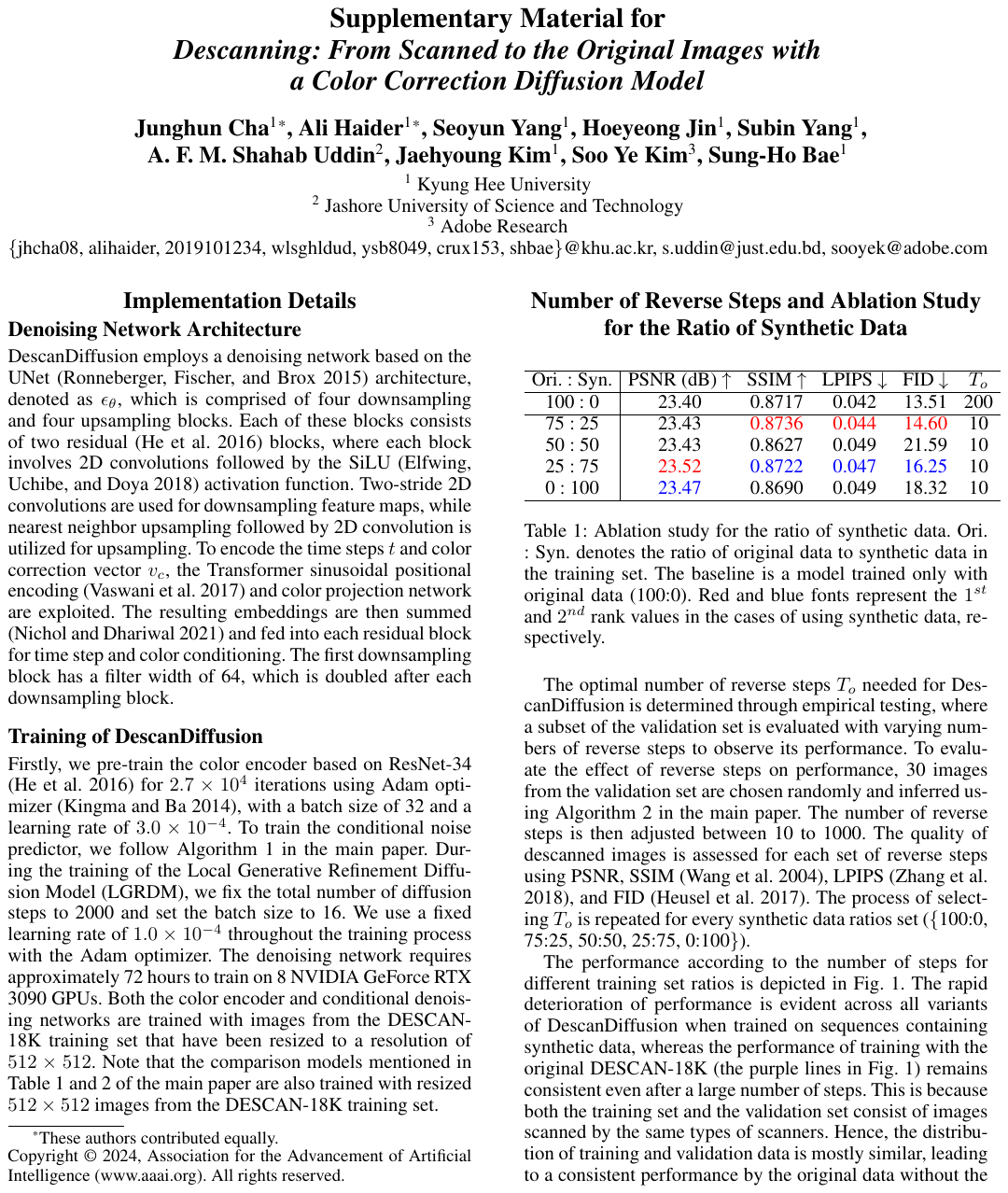}

\end{document}